\documentclass{article} %
\usepackage{iclr2025_conference,times}
\usepackage{comment}  %

\usepackage{amsmath,amsfonts,bm}

\def\eqref#1{equation~\ref{#1}}

\def\1{\bm{1}}

\DeclareMathAlphabet{\mathsfit}{\encodingdefault}{\sfdefault}{m}{sl}
\SetMathAlphabet{\mathsfit}{bold}{\encodingdefault}{\sfdefault}{bx}{n}

\usepackage{hyperref}
\usepackage{url}
\usepackage{subcaption}
\usepackage{graphicx}
\usepackage{algorithmic}  %
\usepackage{float}  %
\usepackage{listings}

\definecolor{codebg}{rgb}{0.95,0.95,0.95}
\definecolor{codeblue}{rgb}{0,0,0.8}
\definecolor{codegreen}{rgb}{0,0.6,0}
\definecolor{codegray}{rgb}{0.5,0.5,0.5}
\definecolor{codepurple}{rgb}{0.58,0,0.82}

\lstdefinestyle{mypython}{
  language=Python,
  basicstyle=\linespread{0.9}\ttfamily\footnotesize,
  commentstyle=\color{codegreen}\itshape,
  keywordstyle=\color{codeblue}\bfseries,
  stringstyle=\color{codepurple},
  backgroundcolor=\color{codebg},
  showstringspaces=false,
  frame=single,        %
  rulecolor=\color{codegray},
  tabsize=4,
  breaklines=true,
  aboveskip=5pt,
  belowskip=5pt,
}

\iclrfinalcopy

\title{Superalignment with Dynamic Human Values}

\author{
  Florian Mai$^{1,3}$ \quad
  David Kaczér$^{1,3}$ \quad
  Nicholas Kluge Corrêa$^{2}$ \quad
  Lucie Flek$^{1,3}$\\[6pt]
  \small
  $^1$\,b-it, University of Bonn \quad
  $^2$\,CST, University of Bonn \quad
  $^3$\,Lamarr Institute for ML and AI\\
  \texttt{\{fmai,dkaczer,kluge,lflek\}@uni-bonn.de}
}

\begin{document}

\maketitle

\begin{abstract}
Two core challenges of alignment are 1) scalable oversight and 2) accounting for the dynamic nature of human values. While solutions like recursive reward modeling address 1), they do not simultaneously account for 2).
We sketch a roadmap for a novel algorithmic framework that trains a superhuman reasoning model to decompose complex tasks into subtasks that are still amenable to human-level guidance. 
Our approach relies on what we call the \emph{part-to-complete generalization} hypothesis, 
which states that the alignment of subtask solutions generalizes to the alignment of complete solutions. 
We advocate for the need to measure this generalization and propose ways to improve it in the future.
\end{abstract}

\section{Introduction}

Alignment of artificial intelligence systems with human values represents
one of the most critical challenges in AI development.
Various techniques have been developed to align AIs with human values~\citep{ji2023ai}, 
with many approaches leveraging human feedback as a key mechanism to judge AI behavior and outputs~\citep{ouyang2022training}. 
Although far from perfect, these human-feedback-based techniques have proven effective 
in scenarios where the tasks performed by the AI remain at or below human intelligence and are relatively low risk, e.g. writing a summary.

However, as AIs eventually move beyond human intelligence, their solutions will become too complex for humans to judge correctly or efficiently, rendering these techniques ineffective.
This is the so-called \emph{scalable oversight problem}\footnote{Scalable oversight is often defined in more general terms as the difficulty of providing human oversight efficiently. In this paper, we use the definition involving superintelligence, also known as \emph{superalignment}.}~\citep{cao2024towards}: \emph{How do you align a superhuman AI with human values?}
This problem becomes especially important as AI agents are performing increasingly higher risk tasks with consequences in the real world.
Promising approaches to scalable alignment leverage the idea of recursive reward modeling, where aligned weak AIs are used to 
align stronger AIs~\citep{leike2018scalable}.
However, although the initial alignment is done by humans, as this approach passes over the threshold of human intelligence, 
the human is removed from the loop entirely: All future alignment is done by AI models without human intervention. 
As \citet{shen2024towards} argue, alignment cannot be a static process, as human values tend to change over time.
In order to preserve human agency, it is therefore necessary to keep humans in the loop, which existing approaches do not account for.

In order to address this issue, we argue that the alignment algorithm must be 
built with inductive biases that keep humans (or human-level AI as a proxy) at the core of the alignment algorithm.
In this paper, we present a roadmap for developing such an approach.
First, we sketch our proposed algorithm, which, inspired by Iterated Amplification~\citep{christiano2018supervising},
decomposes complex tasks into subtasks.
By training a reasoning-based AI to create subtasks simple enough for an aligned human-level AI to solve and judge,
we can ensure that the resulting partial solutions are aligned to human values.
While in contrast to Iterated Amplification this approach removes the need for decomposition by humans,
it introduces the new assumption that the recomposition of aligned partial solutions from subtasks generalizes to an aligned complete solution (illustrated in Figure~\ref{fig:reservation-example}),
which we term the \emph{part-to-complete generalization} hypothesis.
Similarly to other types of alignment generalizations (e.g., weak-to-strong generalization~\citep{burns2023weak}), we argue that this 
assumption is likely to hold to some extent, but needs empirical validation and strengthening through algorithmic innovations.

\begin{figure*}
    \centering
    \begin{minipage}{\textwidth}
        \small

        \noindent
        \begin{minipage}{0.48\textwidth}
        \textbf{Well-aligned sub-tasks:}
                \vspace{-2px}
        \begin{lstlisting}[style=mypython]
# Ask a single user for 
# their preferred restaurants
def ask_preference(user):

# Find a restaurant that satisfies
# all users' preferences
def identify_overlap(preferences):

# Confirm a reservation at the 
# given restaurant
def book_table(restaurant):
        \end{lstlisting}
        \end{minipage}
        \hfill
        \begin{minipage}{0.45\textwidth}
        \textbf{Aligned composition:}
        \vspace{-2px}
        \begin{lstlisting}[style=mypython]
prefs = {}
for u in users:
    prefs[u] = ask_preference(u)
r = identify_overlap(prefs)
book_table(r)
        \end{lstlisting}

        \textbf{Unaligned composition:}
            \vspace{-2px}
        \begin{lstlisting}[style=mypython]
bookings = {}
for u in users:
    prefs = ask_preference(u)
    for r in prefs:
        book_table(r)
r = identify_overlap(prefs)
        \end{lstlisting}
        \end{minipage}
    \end{minipage}
    \caption{Example of part-to-complete generalization in the dinner table reservation task, in which an AI agent is tasked to book a restaurant that satisfies the preferences of all attendees. Partial solutions to sub-tasks are assumed to be well-aligned in isolation. However, the alignment of the complete solution depends on how the partial solutions are recomposed: While in the aligned composition the AI agent first identifies the overlap before booking a single restaurant, in the unsafe composition, tables are booked individually before identifying an overlap, leading to many unnecessary reservations. In Section~\ref{sec:improving-part-to-complete-generalization} we discuss strategies to steer the model toward aligned compositions.}
    \label{fig:reservation-example}
\end{figure*}

\section{Background}

\textbf{Scalable Oversight}
The problem of scalable oversight of deep learning systems
has been identified as a major problem in AI safety~\citep{amodei2016concrete} long before large language models. 
In AI debate~\citep{irving2018ai}, (superhuman) models play a zero-sum debate game to convince a human judge 
that their evaluation of an outcome is better, relying on the assumption that it is easier to convince the judge with true arguments. 
Iterated Amplification (IA)~\citep{christiano2018supervising} relies on humans' ability to safely decompose a complex task into smaller problems that can independently be solved by weak AIs.
While AI debate keeps humans in the loop during alignment, a distinct advantage of IA is
that it constructs strong AIs directly with integrated alignment, lowering the chances of accidents or misuse.
However, human task decomposition is hard to scale to complex tasks, necessitating novel solutions.
Recursive reward modeling~\citep{leike2018scalable}, which describes a family of techniques where a weaker AI model is used
to assist a user in providing feedback to a stronger AI model for training, removes humans from the loop entirely after the first iteration.
In contrast to previous techniques, our approach aspires to both keep humans in the loop and construct strong AIs directly with integrated alignment.

\textbf{Reasoning Models}
With the advent of OpenAI's o1~\citep{jaech2024openai}, 
reasoning models have recently emerged as a new paradigm for training AIs that can solve complex tasks.
Although the exact mechanisms behind o1 are unknown, 
DeepSeek-R1~\citep{guo2025deepseek} is assumed to be the first reproduction of o1. The model leverages a verifier on the generated solution to obtain a quality signal for training via reinforcement learning.
With only a modest amount of training data, DeepSeek-R1 learns to deploy a variety of reasoning capabilities, 
including planning, self-reflection, and self-correction.
This adds to the existing evidence that it is feasible to train reasoning models that learn to decompose tasks into subtasks that can be
solved by a human-level AI~\citep{wen-etal-2024-learning}.

\textbf{Alignment Generalization}
Recent work has explored various forms of generalization in AI alignment. 
\citet{burns2023weak} demonstrate that aligning a strong AI with a weak AI reduces harmfulness while maintaining some capabilities, establishing the concept of weak-to-strong generalization.
However, \citet{shin2024weaktostronggeneralizationdatacentriclens} find that this relies on training examples with both easy and hard predictive patterns.
In parallel, \citet{sun2024easy} observe a similar phenomenon with easy-to-hard generalization, where models trained on simple examples maintain alignment properties when tackling more complex examples.
These different forms of generalization suggest that alignment properties can transfer across capability and complexity levels.
A conceptual limitation of these approaches is their assumption that the weak supervisor is able to
provide non-trivial feedback on some examples, which may not always be the case.
In contrast, our approach trains a stronger AI to directly break down the task into subtasks that are easy for a human-level AI to judge.

\section{Proposed Framework}\label{sec:framework}

Inspired by IA~\citep{christiano2018supervising}, our approach solves harder tasks through task decomposition, but addresses the scalability issues of IA. 
It assumes the existence of aligned human-level AIs and of a correctness verifier.
Figure~\ref{fig:amplifications} shows the algorithm in pseudo-code.

Our approach directly addresses two key challenges of alignment:
1) \emph{Scalable oversight}: By decomposing them into subtask structures of respective complexity, this model is able to solve increasingly complex tasks (Figure~\ref{fig:algorithm}) while still producing solutions that are aligned to human values.
2) \emph{Dynamic nature of human values:} By enforcing that each subtask is solved by a human-level AI, we can incorporate evolving human values by continuously updating the human-level AI proxy accordingly, e.g., through RLHF~\citep{christiano2017deep} (see Figure~\ref{fig:superalignment}).

\section{Measuring Part-to-Complete Generalization}\label{sec:measuring-part-to-complete-generalization}

\begin{figure*}[h]
   \centering
   \begin{subfigure}[b]{0.45\textwidth}
       \centering
       \fbox{\begin{minipage}{1.0\textwidth}
       \small
       \textbf{Inputs:} Human H, Human-level AI $H_{\phi}$, Planner $P_{\theta}$, Dataset $\mathcal{D}$ of hard superhuman tasks $X_i$, Dataset $\mathcal{E}$ of human-level tasks $Y_i$, Verifier $V$
   \vspace{0.15cm}

       \textbf{procedure} scalable\_align($P_\theta, H_{\phi}, H, \mathcal{D}, \mathcal{E}$, V):

       1. \textbf{while} True \textbf{do} \\
       2. \hspace{1em} $H_{\phi} \gets \text{align\_to\_human}(H, H_{\phi}, \mathcal{E})$\\
       3. \hspace{1em} \textbf{for} $X_i$ in $\mathcal{D}$ \textbf{do}\\
       4. \hspace{3em} $P_\theta \gets \text{train\_planner}(P_\theta, H_{\phi}, X_i, V)$\\ [0.5em]
       \textbf{procedure} align\_to\_human($H, H_{\phi}, \mathcal{E}$):\\
       1. \textbf{for} $Y_i$ in $\mathcal{E}$ \textbf{do}\\
       2. \hspace{2em} $H_{\phi} \gets align(H_{\phi}, H, Y_i$) \\
       3. \textbf{return} $H_{\phi}$
       \end{minipage}}
       \caption{Alignment to evolving human values.}
       \label{fig:superalignment}
   \end{subfigure}\hfill
   \begin{subfigure}[b]{0.53\textwidth}
       \centering
       \fbox{\begin{minipage}{0.95\textwidth}
       \small
       \textbf{procedure} train\_planner($P_\theta, H_{\phi}, X, V$):\\
       1. $subtasks \gets P_\theta.\text{decompose}(X)$\\
       2. $partial\_solutions \gets []$\\
       3. $partial\_rewards \gets []$\\
       4. \textbf{for} $t$ in $subtasks$ \textbf{do}:\\
       5. \hspace{2em} $s \gets H_{\phi}.\text{solve}(t)$\\
       6. \hspace{2em} $r \gets H_{\phi}.\text{judge}(s)$\\
       7. \hspace{2em} $partial\_solutions.\text{append}(s)$\\
       8. \hspace{2em} $partial\_rewards.\text{append}(r)$\\
       9. $S \gets P_\theta.\text{recompose}(partial\_solutions)$\\
       10. $R \gets V.\text{verify}(X, S)$\\
       11. $P_\theta \gets \text{RLFT}(P_\theta, R + \text{sum}(partial\_rewards))$ \\
       12. \textbf{return} $P_\theta$

       \end{minipage}}
       \caption{Training the planner via task decomposition.}
       \label{fig:algorithm}
   \end{subfigure}
   \caption{Our proposed approach (see Section~\ref{sec:framework}) for maintaining human oversight in superalignment through part-to-complete generalization.
   (a) On a regular basis, a human-level AI $H_{\phi}$ is aligned to humans $H$ on human-level tasks $\mathcal{E}$ to account for the dynamic nature of human values.
   After adapting the human-level AI, we train the superhuman planner model $P_{\theta}$ on superhuman tasks $\mathcal{D}$.
   (b) A reasoning model $P_\theta$ decomposes each task $X$ into simpler subtasks. Each subtask is solved and judged by the human-level aligned AI $H_{\phi}$. The reasoning model then recomposes the partial solutions into a complete solution, which is verified for correctness using a rules-based verifier $V$. The reasoning model is then updated using a reinforcement learning algorithm RLFT (e.g., PPO~\citep{schulman2017proximal}) based on the correctness reward $R$ and partial alignment rewards.
   With the \emph{part-to-complete generalization} hypothesis, we expect the alignment of solutions to subtasks to generalize to the complete solution. 
   }
   \label{fig:amplifications}
\end{figure*}

Our framework targets tasks that a human-level AI is not able to judge the complete solution of, 
but whose AI-generated partial solutions can be judged reasonably well 
by a human-level AI\footnote{We assume that the human-level AI is able to make value-consistent judgements across partial solutions.}, which is used as alignment feedback for the reasoning model.
However, the alignment of partial solutions does not necessarily generalize to the alignment of the complete solution.
Figure~\ref{fig:reservation-example} illustrates this challenge with a simple example of an AI agent booking a dinner table.

Our approach assumes what we term \emph{part-to-complete generalization}, where despite a lack of feedback on complete
solutions, the reasoning model learns to generate compositions of partial solutions that are still aligned with human values.
Analogous to the suspected mechanism in weak-to-strong generalization~\citep{burns2023weak}, we hypothesize that the AI understands the intent of alignment of partial solutions.
However, the extent to which this holds is an empirical question that is not trivial to answer; future research needs to study
the extent of part-to-complete generalization for different types of tasks and reasoning models. %
Using the sandwiching method for scalable oversight~\citep{bowman2022measuring}, we can evaluate the performance of our approach on risk-laden agentic domain-expert tasks when receiving feedback on partial solutions from a non-expert~\citep{zhou2024haicosystem}.

\section{Improving Part-to-Complete Generalization}\label{sec:improving-part-to-complete-generalization}

Similarly to how \citet{burns2023weak} propose methods that improve weak-to-strong generalization, we expect that there are several ways to improve the part-to-complete generalization of reasoning models.

\textbf{(1) Restricted Composition Space}
One approach to improve part-to-complete generalization is to restrict the space of possible compositions. By limiting compositions to specific structures like trees or sequential processes, we can eliminate certain classes of safety violations by design. For example, in our example of dinner reservation (Figure~\ref{fig:reservation-example}), limiting the depth of nested for-loops to one could have prevented the unaligned complete solution. 

\textbf{(2) Balanced Subtask Complexity}
Another strategy is to carefully balance the complexity of generated subtasks. 
The planner should aim to create subtasks that require approximately human-level intelligence to understand and verify, 
as simpler subtasks would push more complex logic into the composition step itself. 
This balance ensures that the human-level AI can effectively judge each component while minimizing the added complexity that 
emerges during composition.

\textbf{(3) Solution Summarization}
A third approach involves generating high-level summaries of complete solutions for human-level AI judgment. 
While the human-level AI may not comprehend all implementation details, 
it can provide an imperfect evaluation of whether the summary adequately reflects the intended composition of subtasks and maintains desired safety properties. 
Despite the imperfect evaluation, the planner model may learn the true intent through weak-to-strong generalization.

\section{Discussion and Conclusion}\label{sec:discussion}
In their review of the alignment literature, \citet{shen2024towards} list concrete challenges of alignment that a promising roadmap for alignment should address.
In this section, we discuss our framework in light of these challenges.

By design, our framework directly addresses the challenge of \textbf{scalable oversight}. 
Rather than attempting to verify increasingly complex behaviors as a whole, we maintain oversight by ensuring that 
all solutions are composed of aligned partial solutions. 
This approach scales naturally with AI capability: As the planner becomes more sophisticated, 
it can create more complex de- and recompositions of hard tasks while keeping individual subtasks at human-level difficulty.
Moreover, our approach increases robustness to \textbf{specification gaming} through multiple layers of oversight. 
By decomposing complex tasks into human-verifiable subtasks, we make it harder for the system to find and exploit loopholes, 
as each component must pass human-level AI verification. 
Furthermore, the part-to-complete generalization property ensures that gaming the specification at the composition level 
would require simultaneously satisfying multiple independent human-aligned constraints, 
making unintended solutions less likely.
In contrast to previous scalable oversight solutions, our framework is equipped to account for the \textbf{dynamic nature of human values}. 
Since oversight is maintained through a human-level AI proxy, updates to human values can be incorporated by updating this 
proxy, which then influences both the verification of subtasks and the training of the planner. 
This creates a dynamic feedback loop where changes in human values naturally propagate through the system without 
requiring complete retraining.
Finally, our framework provides several safeguards against \textbf{existential risk}. 
First, capability and alignment are developed simultaneously rather than sequentially, preventing unaligned superhuman AI to be developed in the first place. 
Second, the decomposition approach ensures that any potentially dangerous capabilities must be constructed from human-verified components, 
making it harder to develop harmful behaviors unnoticed.

Although our framework addresses many key alignment challenges, significant work remains. 
While it addresses the outer-alignment problem, it does not directly address the alignment of a model’s internal objectives (inner alignment).
Moreover, its success hinges on the extent of part-to-complete generalization, which must be empirically validated across different domains and task complexities. New methods to improve part-to-complete generalization need to be developed.
These challenges, while substantial, represent concrete research directions rather than fundamental limitations of the approach, and we invite the community to join us in addressing them.

\subsubsection*{Acknowledgements}
This work has been partially supported by the state of North Rhine-Westphalia as part of the Lamarr Institute for Machine Learning and Artificial Intelligence.
Nicholas Kluge Corrêa is funded by the Ministerium für Wirtschaft, Industrie, Klimaschutz und Energie des Landes Nordrhein-Westfalen (Ministry for Economic Affairs, Industry, Climate Action and Energy of the State of North Rhine- Westphalia), as part of the KI.NRW-flagship project "Zertifizierte KI" (Certified AI).

\bibliography{iclr2025_conference}
\bibliographystyle{iclr2025_conference}

\end{document}